\documentclass{egpubl}
\usepackage{eurovis2021}

\SpecialIssuePaper         

\usepackage[T1]{fontenc}
\usepackage{dfadobe}  
\usepackage{enumitem}
\setlist[enumerate]{
  labelsep=8pt,
  labelindent=0.0\parindent,
  itemindent=0pt,
  leftmargin=2.0\parindent,
  before=\setlength{\listparindent}{-\leftmargin},
}

\usepackage{cite}  
\BibtexOrBiblatex
\electronicVersion
\PrintedOrElectronic
\ifpdf \usepackage[pdftex]{graphicx} \pdfcompresslevel=9
\else \usepackage[dvips]{graphicx} \fi

\usepackage{egweblnk}

\title[VICE]%
      {VICE: Visual Identification and Correction of Neural Circuit Errors}
      
\author[F. Gonda et al.]
{\parbox{\textwidth}{\centering Felix Gonda$^{1}$\orcid{0000-0003-1870-0905}, Xueying Wang$^{2}$\orcid{0000-0002-2399-8083}, Johanna Beyer$^{1}$\orcid{0000-0002-3505-9171}, Markus Hadwiger$^{3}$\orcid{0000-0003-1239-4871}, Jeff W. Lichtman$^{2}$, and Hanspeter Pfister$^{1}$\orcid{0000-0002-3620-2582}
        }
        \\
{\parbox{\textwidth}{\centering $^1$Visual Computing Group, Harvard University, Cambridge, Massachusetts, United States\\
         $^3$Visual Computing Center, KAUST, Thuwal, Saudi Arabia\\
         $^2$Department of Molecular and Cellular Biology, Harvard University, Cambridge, Massachusetts, United States
       }
}
}

\begin{document}

\teaser{
 \includegraphics[clip, trim=0.0cm 0.0cm 0.0cm 2.5cm, width=\linewidth]{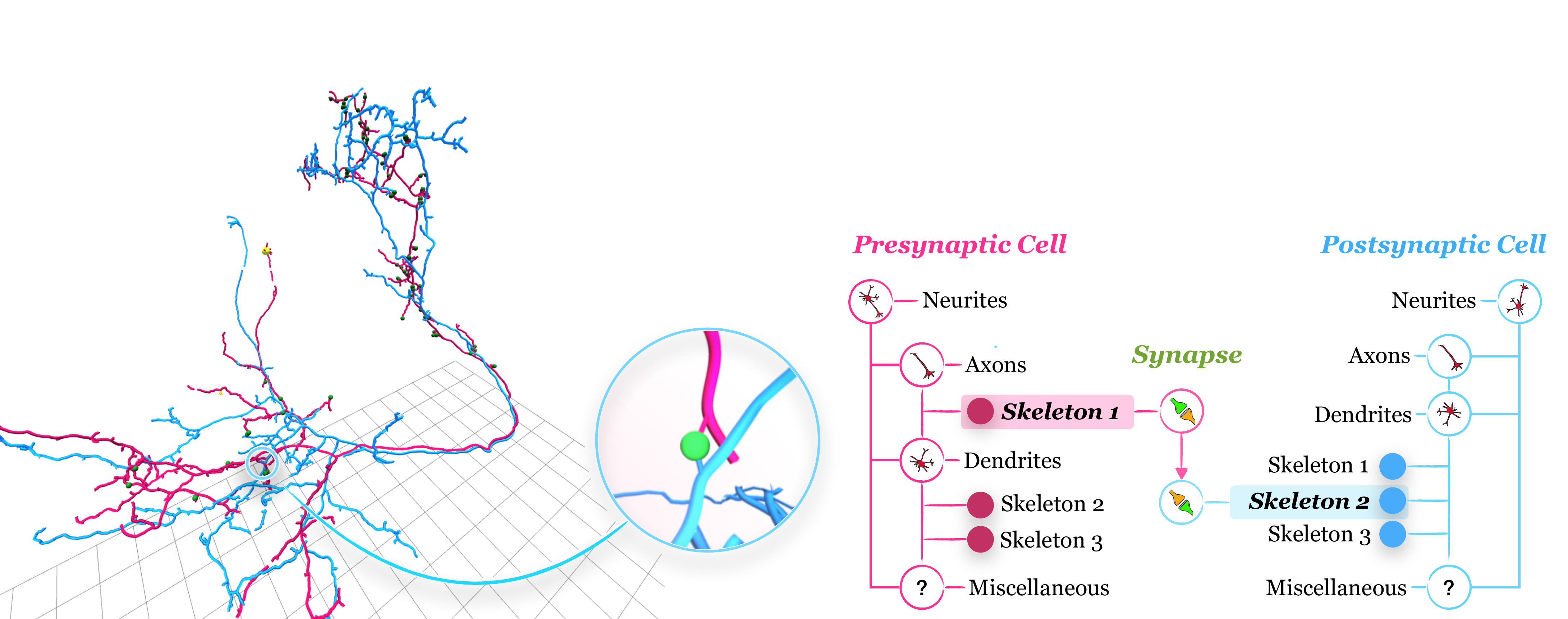}
 \centering
  \caption{A local circuit of a presynaptic cell in targeted proofreading of structures with similar morphological features in the output portion of the Drosophila melanogaster brain \cite{hemibrain}.  VICE unravels connectivity pathways at the level of individual axons and dendrites where the presynaptic and postsynaptic elements of synapses reside.}
\label{fig:teaser}
}

\maketitle
\begin{abstract}
A connectivity graph of neurons at the resolution of single synapses provides scientists with a tool for understanding the nervous system in health and disease. Recent advances in automatic image segmentation and synapse prediction in electron microscopy (EM) datasets of the brain have made reconstructions of neurons possible at the nanometer scale. However, automatic segmentation sometimes struggles to segment large neurons correctly, requiring human effort to proofread its output. General proofreading involves inspecting large volumes to correct segmentation errors at the pixel level, a visually intensive and time-consuming process.  This paper presents the design and implementation of an analytics framework that streamlines proofreading, focusing on connectivity-related errors. We accomplish this with automated likely-error detection and synapse clustering that drives the proofreading effort with highly interactive 3D visualizations. In particular, our strategy centers on proofreading the local circuit of a single cell to ensure a basic level of completeness. We demonstrate our framework's utility with a user study and report quantitative and subjective feedback from our users. Overall, users find the framework more efficient for proofreading, understanding evolving graphs, and sharing error correction strategies.


\end{abstract}  
\section{Introduction}\label{sec:introduction}
Connectomics is a sub-area of Neuroscience that seeks to reconstruct the wiring diagram of the connectivity between individual neurons in the brains of organisms. A complete wiring diagram of a brain, also known as a connectome, is considered by scientists as an essential step of understanding the brain \cite{connectome}.  The information contained in a connectome is critical for biologists to gain insights into the brain's functional structure, as demonstrated in efforts on \textit{Caenorhabditis elegans}~\cite{celagan}, \textit{Drosophila melanogaster} \cite{Takemura13711}, \textit{zebrafish} \cite{zebrafinch} and \textit{mice}~\cite{netanatomy}. 
\begin{figure}[t]
 \centering 
 \includegraphics[clip, trim=3.75cm 0.0cm 3.75cm 0.0cm, width=\columnwidth]{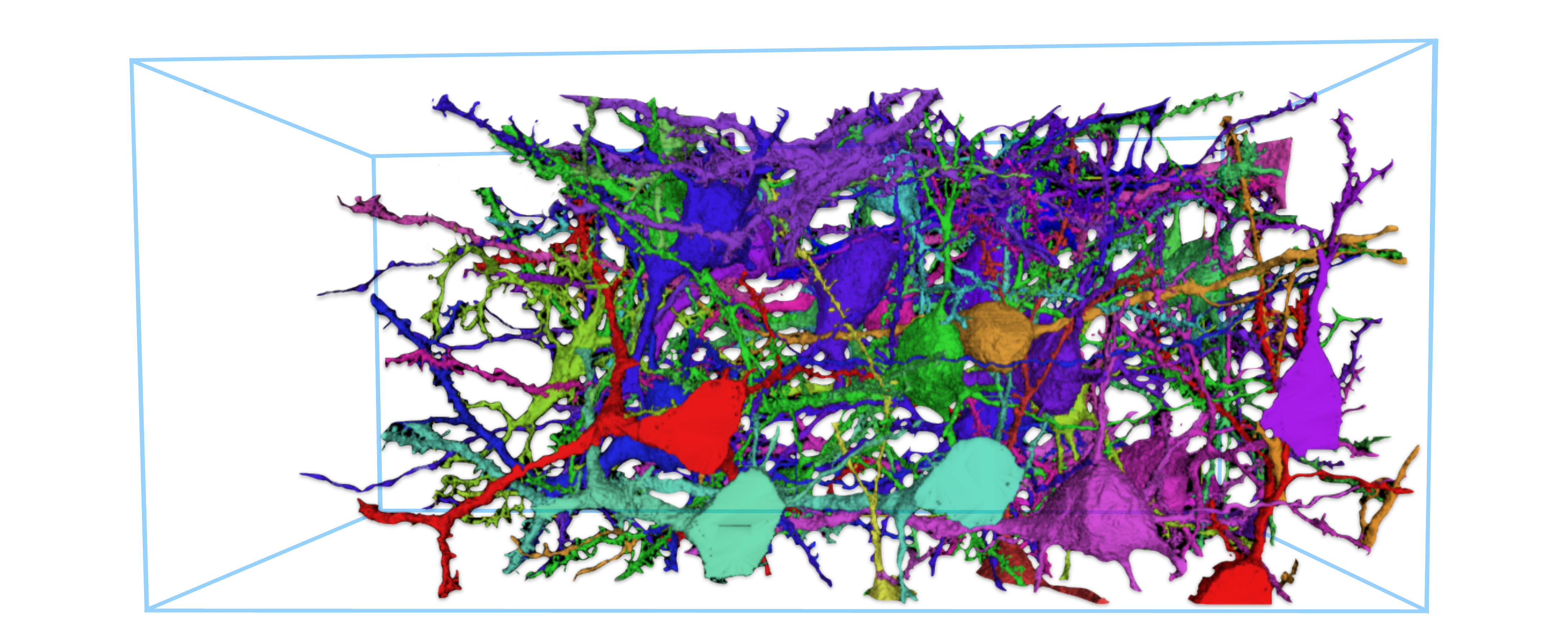}
\caption{A visualization of a Zebra finch brain's reconstruction in Neuroglancer~\cite{neuroglancer}.  The density of cells makes it difficult to discern connectivity between cells.}
\vspace{-10pt}
\label{fig:neuroglancer_vis}
\end{figure}

However, reconstructing a connectome from high-resolution EM images of brain tissue requires extensive human effort to proofread automatically generated segmentation and synapses.  Recent advancements in EM imaging \cite{richard2016imaging}, segmentation~\cite{floodfilling,Funke2019LargeSI}, and synapse prediction~\cite{synapsepred} create new challenges to accelerate proofreading. Biologists can now generate reconstructions from large datasets (in multi-terabytes) where the segmentation is largely correct, the morphological structures of neurons can be very large, and the density of synapses can be high. At this scale, proofreading by human experts becomes infeasible. For example, in the \textit{Drosophila melanogaster} brain \cite{hemibrain}, synapse prediction produces about 9 million presynaptic contacts and 60 million post-synaptic-densities.  It would take a trained person 230 working years to manually validate each site at a rate of 1000 connections per day. Furthermore, the density of neurons and synapses in neural tissue pose great difficulties in presenting the connectome in a way that is conducive to analysis. A simple visualization of reconstruction in 3D, such as in the Zebra finch brain in Figure \ref{fig:neuroglancer_vis}, quickly clutters the view and makes it challenging to discern connectivity.

This paper proposes a semi-automated approach to address the proofreading and visualization of neural connectivity graphs jointly. In most state-of-the-art methods, neurons are proofread in parallel at the same time, usually by proofreading the volume slice by slice. Our approach, which is based on dialog with neuroscientists, employs a single-cell strategy to capture connectivity pathways. In this strategy, proofreading begins with the cellular compartments of a presynaptic cell, as shown in Figure~\ref{fig:teaser}, and progresses outward to capture its entire local circuit.  This strategy has several advantages over slice-based proofreading.  First, it ensures a basic level of completeness of each cell and enables scientists to perform analysis tasks, such as the classification of cellular components, on completed sub-graphs.  Second, the strategy significantly reduces the number of neurons to the single-cell and its immediate partners, making it easier to visualize morphology and connections jointly without clutter. Finally, the strategy reduces the memory and speed requirements for visualizing large neurons at interactive rates. We realized the single-cell strategy with a visual analytics framework that detects likely connectivity errors in reconstructions and aggregates co-located synapses into clusters to guide the proofreading effort.  To interface with the framework, we design highly interactive visualizations and editing tools to enable fast error correction and synapse validation.  This approach enables us to accelerate proofreading and present connectivity graphs in an uncluttered and conducive way for proofreading and analysis.

We make the following contributions. Our first contribution is \emph{an automatic error detection system} for identifying likely connectivity errors in automatic neuron reconstructions.  This error detection system, based on heuristics elicited from domain experts, drives the proofreading effort.  Our second contribution is the \emph{VICE framework's design} and implementation that manifests our single-cell proofreading strategy with a top-down 3D visualization of a connectivity graph, guided by the error detector.  To enable rapid examination of synapse sites and fast correction of errors, we design \emph{a scalable 3D inspection and editing tool} that attaches to cellular structures as our third contribution. This tool enables proofreading of large volumes by supporting on-demand loading of segmentation and image data within a small region surrounding an attachment point. For our final contribution, we conduct \emph{a user study to assess our proposed method's efficiency} and usability compared to the previous proofreading approach of our collaborators.  
\section{Related Work}\label{sec:related_work}
In this section, we describe segmentation and visualization works in connectomics that are related to our work.

\noindent
\textbf{Segmentation for Connectomics}\\
In connectomics, segmentation involves processing EM data to extract the morphology of individual neurons and the synapses between them.  Automatic algorithms such as Flood Filling Networks \cite{floodfilling} have been shown to perform the segmentation task well \cite{hemibrain}. However, irrespective of data size, current segmentation algorithms generalize poorly in regions with low contrast, low resolution, or image artifacts.  This is especially the case for small neurites where synapses often reside \cite{AnalyzingImage}.  Addressing this issue typically requires extensive manual proofreading by human experts. The most common approaches to proofreading utilize desktop applications such as VAST \cite{vast}, collaborative tools such as CATMAID~\cite{catmaid}, Dojo~\cite{dojo}, EyeWire~\cite{eyewire}, and Neutu~\cite{neutu}, and active learning approaches \cite{guidedproofreading}. These tools operate on dense segmentation and fix errors either at the pixel level on a slice-by-slice basis (e.g., VAST, Dojo), object-level (e.g., Neutu), or by separating data into smaller blocks and proofreading each block individually (e.g., EyeWire). In the active learning approach~\cite{guidedproofreading}, a machine learning algorithm is used to detect merge and split errors in segmentation.  The errors are then corrected by prompting the user to make binary decisions. The amount of manual human effort required to correct errors in these tools remains an issue as the size of data increases.  

Unlike the general proofreading approaches that focus on correcting errors to reconstruct neurons' anatomical structures, our approach focuses on correcting and validating connectivity pathways. This approach is advantageous to our collaborators because it enables them to trace the pathways of a single cell and perform analysis on sub-graphs.  As such, we detect likely connectivity errors in segmentation as a first step and use these error locations to drive proofreading. Unlike the guided proofreading approach of Haehn \textit{et al.}~\cite{guidedproofreading} which detects segment errors using a boundary detector and prompt users for binary decisions, our approach is based on heuristics derived from actual proofreading scenarios and generates likely connectivity errors by taking into account predicted synapse locations and does not prompt users for decisions. 

\noindent
\textbf{Visualization for Connectomics.}\\
In connectomics, visualization focuses on the exploration of segmented data and analysis of neuronal connectivity\cite{daniel2017scalable}.  An overview of existing tools for visualization in connectomics is given by Pfister \textit{et al.}~\cite{visconnectomics}, covering multiple scales of connectivity.  Current visualization techniques in connectomics either follow 2D, 3D, or a hybrid approach.  2D visualization is typically employed for dense pixel-level proofreading when it is necessary to verify if some inconsistency is caused by reconstruction errors or random structures in the EM data\cite{vast}.  3D visualization is employed for neuron morphology, such as in Neuroglancer~\cite{neuroglancer}, and ConnectomeExplorer~\cite{ConnectomeExplorer}.  Neuroglancer provides a web-based viewer for volumetric data with support for arbitrary cross-sectional views without connectivity.  For large-scale exploration, ConnectomeExplorer provides a sophisticated query system for analyzing objects on multiple domains. Connectivity, in ConnectomeExplorer, is limited to a node-based abstract graph representation. An approach that combines structure and connectivity exploration at the synapse level is NeuroLines \cite{neurolines}. In NeuroLines, structure and connectivity information is abstracted to a 2D representation using a visual subway map metaphor that honors relative distances between structures and branches to preserve topological correctness. A more recent work, neuPrint \cite{neuprint}, organizes a connectome's data into a repository and provides connectivity visualization with an adjacency matrix.  A complementary approach to our work is NeuroBlocks~\cite{awami_neuroblocks_2015}, which provides auditing and provenance management of segmentation data to help scientists manage a large proofreading project. NeuroBlocks provides a pixel view that gives an overview of the proofreading progress of structures in terms of completion status and date. However, NeuroBlocks itself does not perform proofreading; it provides an API for connecting external applications.  Therefore, even if our framework were integrated into NeuroBlocks, our collaborators would still have to switch between slice-based proofreading in VAST and separate tools for 3D visualization and connectivity exploration.

In our approach, we visualize local connectivity graphs of neurons in 3D. This approach is critical for our collaborators as it enables them to scrutinize connectivity pathways and rapidly resolve errors in a global context.  

\section{Design Process}\label{sec:motivation}
Our design process consists of working with domain experts from the Center for Brain Science at Harvard University.  For 20 weeks, we engaged in a user-centered iterative design process where we met with domain experts on a bi-weekly basis. Our meetings typically lasted 30 minutes.  The first eight weeks were dedicated to requirements gathering and prototyping, while the remaining weeks were used to show updates and collect feedback from experts. Requirements gathering consisted of semi-structured interviews that were conducted at a proofreader's office.  The interviews consisted of demonstrations of actual proofreading scenarios and discussion of bottlenecks. Some sessions involved eliciting expert heuristics for identifying errors that could be automated to accelerate proofreading. For design iterations, we first demonstrated the most recent updates and asked experts to test the system and provide feedback. We used the feedback from these sessions to refine the design and address problems.

\subsection{Domain Experts}\label{sec:domain_experts}
Our domain experts consist of one post-doctoral researcher and six proofreaders (four undergraduate students and two graduate students).  The researcher provides data and biological knowledge for neuron and synapse reconstruction.  Three students are trained to proofread morphology reconstruction, and three are trained on synapses validation.  All proofreaders were recruited and trained by the researcher.

\begin{figure*}[t!]
\begin{minipage}[b]{1.0\linewidth}
  \centering \centerline{\includegraphics[clip, trim=0.0cm 8.5cm 0.0cm 7.5cm, width=\textwidth]{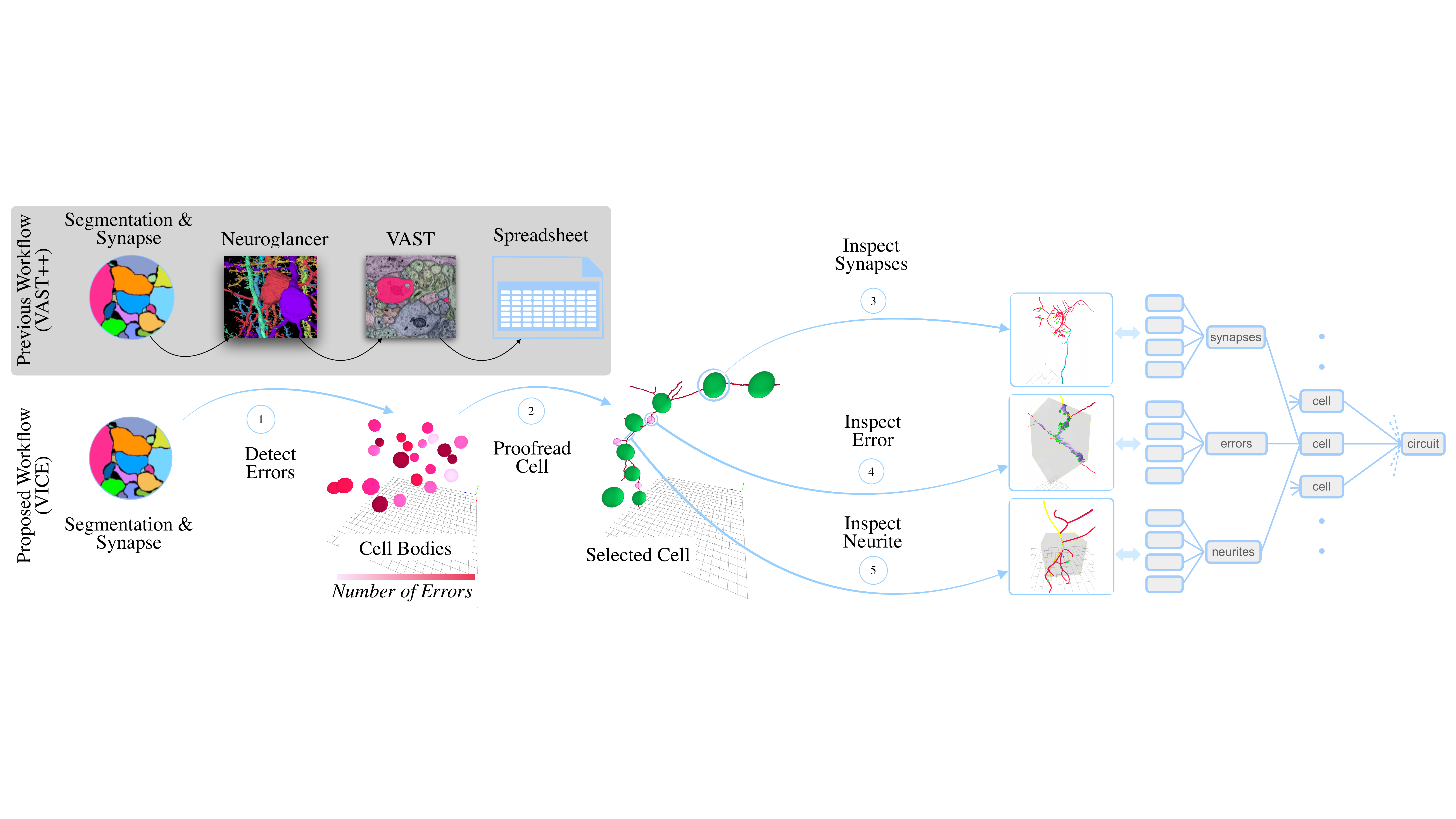}}
\end{minipage}
\caption{\label{fig:workflow}
Compared to the previous workflow (gray), our approach consists of five steps: (1) error detection, (2) single cell proofreading initiation, (3) synapse inspection and validation, (4) error inspection and correction, and (5) neurite inspection. The previous workflow runs three applications simultaneously to manually proofread reconstructions.
}
\vspace{-10pt}
\end{figure*}

\subsection{Goals}\label{sec:goals}
This work's motivation stems from dialog with domain experts, including one of the authors, on the challenges of proofreading connectome reconstructions. When working with a new reconstruction, our collaborators often explore the data in Neuroglancer to identify suspicious structures. This process involves scrolling through layers of image data and peering at predicted structures from multiple angles to determine errors.  Coordinates of suspicious structures are then manually transferred to a separate tool, VAST, for dense correction. For synapse validation, each site is visually inspected in VAST and documented in a spreadsheet. At a large scale, this process becomes infeasible.

Our interaction with domain experts also included an investigation into the precise nature of the data used in proofreading, its scale, and the properties necessary for constructing connectivity graphs in 3D.  An immediate concern was that early experiments in standard 3D graph layout tools such as graphs with 3D force-directed layouts~\cite{3dforcegraph} in Three.Js~\cite{threejs} had produced poor results due to the density of nodes. We initially visualized connections between cells to investigate the problems with 3D force graphs but quickly abandoned the layout as it led to illegible hairballs.  Visualizing neurons in Neuroglancer led to a cluttered view that obscured connectivity between cells, as depicted in Figure \ref{fig:neuroglancer_vis}. 
These experiments pointed to several challenges that make proofreading and visualizing connectivity graphs problematic.
\begin{enumerate}[label=\textbf{C\arabic*}]
\item The \textbf{density and complexity} of neurons in EM reconstructions makes it challenging to visualize neurons in 3D space simultaneously without clutter.
\item Automatic reconstruction typically generates \textbf{partial neurons}; as such, a single neuron could span multiple volumes before it is fully reconstructed.
\item Most reconstructed neurons have a \textbf{large number of synapses}, making it difficult to visualize them without clutter. Furthermore, automatic reconstruction methods typically generate spatially unsorted synapses, leading to the redundant inspection of image data.
\item Managing the \textbf{sheer amount of EM data}, including raw image data, segmentation data, and metadata, is challenging and requires special consideration when creating interactive visualizations.
\item The use of \textbf{multiple applications} to carry out proofreading tasks demands greater concentration when from users, leading to unintended errors.
\end{enumerate}


\subsection{Domain Specific Tasks}\label{sec:task_analysis}
Motivated by an apparent need for a tool to address the challenges described in Section~\ref{sec:goals}, we worked with our domain experts to identify the key tasks that such a tool should support.
\begin{enumerate}[label=\textbf{ T\arabic*}]
\item \textbf{Identify connectivity errors} in automatic reconstructions.  The locations of these error sites should guide proofreading.
\item \textbf{Optimize synapse validation} by emphasizing the joint-proofreading of co-located synapses along cell branches. This will eliminate redundant inspection of image data. 
\item \textbf{Examine connectivity graphs} at: (a) the global level, in terms of the distribution of cell bodies, synapses, and errors; and (b) the local level in terms of connectivity pathways to immediate partners.  
\item \textbf{Reduce the manual documentation} of cellular components by removing redundant and unnecessary manual inputs.  This improvement will help users to focus on fixing errors.
\item \textbf{Generate an overview} of the reconstruction data to help users navigate cellular compartments.
\end{enumerate}

\section{Visual Identification and Correction of Connectivity Errors}\label{sec:method}
We now describe the design of our analytics framework for visual identification and correction of connectivity errors (VICE), which aims to help users with the tasks defined in Section~\ref{sec:task_analysis}. VICE consists of two parts.  An automatic part that detects errors, clusters synapses, classifies neurites, and an interactive part that visualizes and edits a connectivity graph. The automatic part's output drives the interactive part to enable fast proofreading of problematic areas of the graph with highly interactive 3D interfaces.  We first describe the basic workflow and visual components of our framework.  Then, we describe a sequence of strategies and design choices to address the tasks and challenges described in Section~\ref{sec:task_analysis} and~\ref{sec:goals}. We finally discuss our implementation.

\begin{figure*}[t!]
\begin{minipage}[b]{1.0\linewidth}
  \centering \centerline{\includegraphics[clip, trim=0.5cm 0.0cm 0.5cm 0.0cm, width=\textwidth]{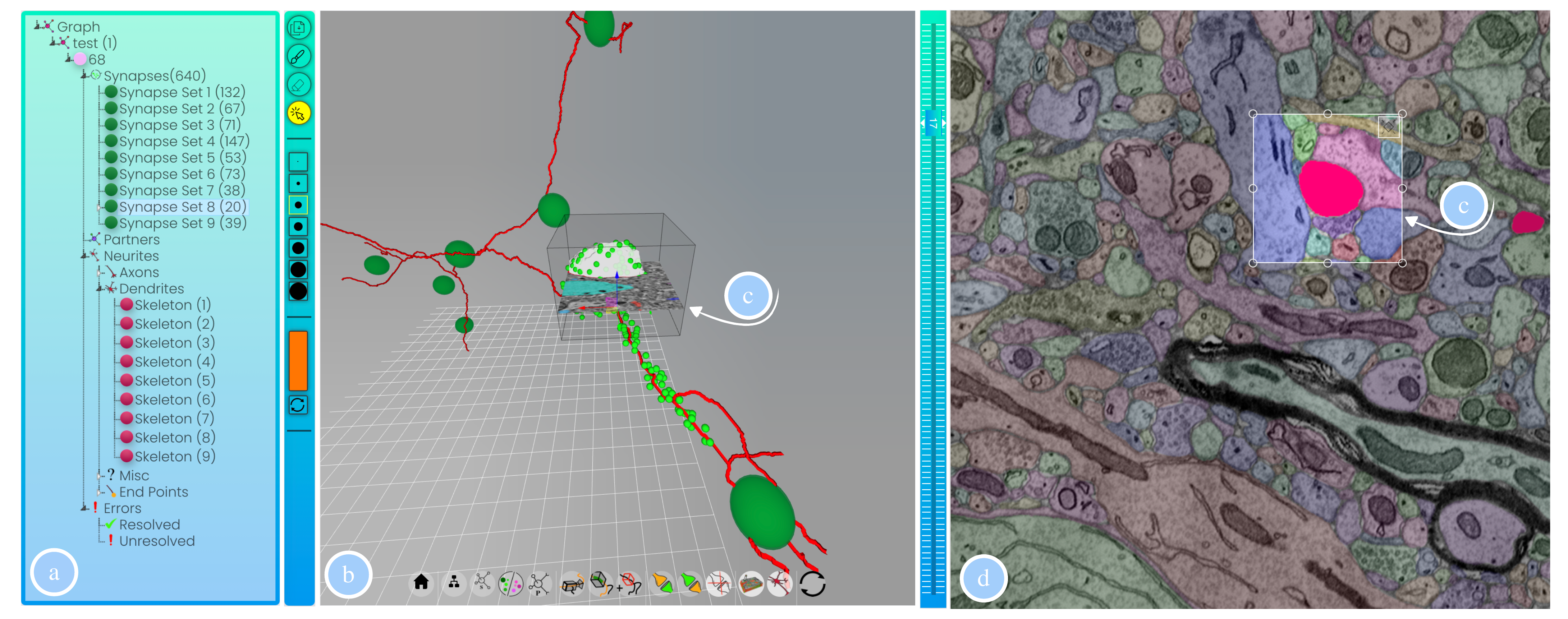}}
\end{minipage}
\caption{\label{fig:userinterface}
VICE's interactive part consists of three visual components. (a) A circuit browser that decomposes a circuit into cellular compartments for rapid exploration and classification.  (b) A 3D viewer that renders the local circuit of a cell. And, (c) an inspector tool for scrutinizing and correcting problematic areas of the circuit and exploring the original EM data. (d) A slice view of the current volume at large scale.
}
\vspace{-10pt}
\end{figure*} 

\subsection{Workflow}\label{sec:workflow}
Figure \ref{fig:workflow} shows a comparison of our proposed workflow to our collaborators' previous workflow for proofreading. Previously, our collaborators ran three applications: VAST, Neuroglancer, and a spreadsheet editor side-by-side, referred to as VAST++.  Proofreading in VAST++ involves iteratively using these applications to manually identify and correct errors, validate synapses, and document cellular structures. Our proposed workflow is structured into five steps. 

\noindent
\textbf{Detect Errors}. In the first step, we detect likely connectivity errors in automatic reconstruction.

\noindent
\textbf{Proofread Cell}. In the second step, a user initiates proofreading by selecting a cell body, shown in red in Figure~\ref{fig:workflow}, to load cellular compartments.

\noindent
\textbf{Inspect Error}. To inspect an error, a user selects its region of interest to launch the inspection and correction tool described in Section~\ref{sec:userinterface}.

\noindent
\textbf{Proofread Synapses}. To inspect synapses, a user selects a synapse cluster to inspect its individual synapses using the inspection and correction tool.

\noindent
\textbf{Proofread Neurite}. To proofread a neurite, the user can select its surface to launch the inspection and correction tool.

\subsection{Visual Components}\label{sec:userinterface}
We design the interactive part of VICE as a single interface following Shneiderman's principle: overview first, zoom and filter, then details on demand~\cite{eyeshaveit}. The interface consists of a neural circuit browser, a viewer, and an inspector (\textbf{C5}). Each of these visual components provides a different level of detail.  The browser allows quick exploration based on a circuit's structural components. The viewer is for exploration of the 3D morphology and spatial distribution of synapses. The inspector is for error correction and detailed exploration of the original EM data. These visual components, shown in Figure~\ref{fig:userinterface}, are synchronized to reflect user choices. 
\vspace{3mm}\\
\noindent
The \textbf{Circuit Browser}\label{sec:circuit_browser}, shown in Figure~\ref{fig:userinterface} (a), is a tree view that represents a hierarchical decomposition of a circuit into cellular compartments (\textbf{T5}). This representation serves several purposes. First, it gives an overview of the reconstructed circuit at a glance, allowing users to quickly navigate cellular compartments and components.  Second, it supports on-demand loading and unloading of cellular compartments, enabling efficient exploration of large reconstructions. Third, it supports documentation and classification of cellular components using bulk operations - for instance, a group of synapses or neurites can be classified jointly by multi-selecting from the browser and applying a label. This bulk operation capability enables users to significantly reduce turnaround time for proofreading (\textbf{T4}). Finally, the browser supports global and local exploration modes. At the global level, users can examine the distribution of cell bodies and the concentration of synapses and errors. At the local level, users can examine connectivity pathways at the cellular level (\textbf{T3}).  
\begin{figure*}[t!]
\begin{minipage}[b]{1.0\linewidth}
  \centering \centerline{\includegraphics[clip, trim=1.0cm 7.0cm 1.0cm 7.0cm, width=\textwidth]{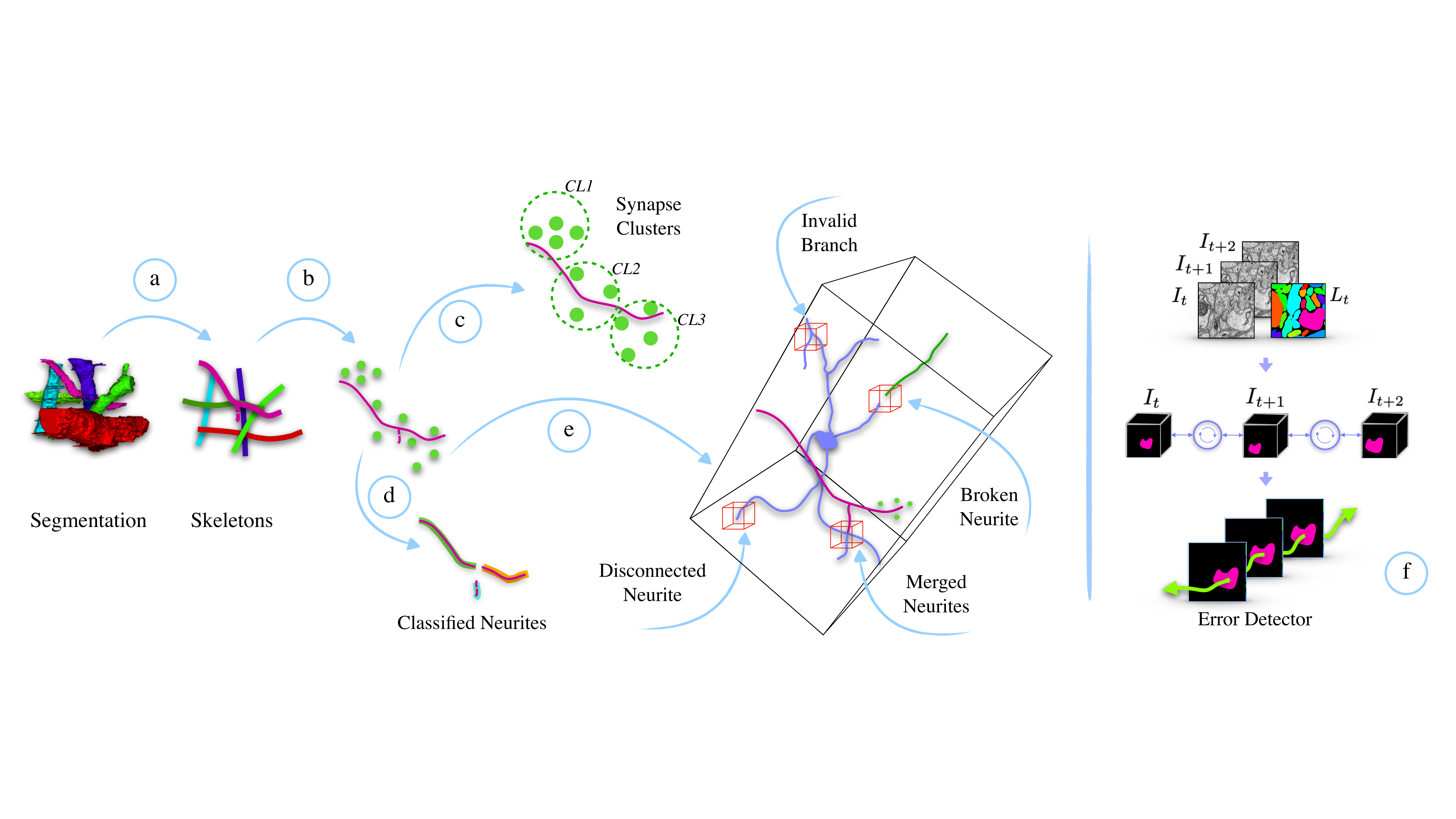}}
\end{minipage}
\caption{\label{fig:strategies}
We accelerate proofreading by automatically:  (a) transforming segmentation into skeletons, (b) associating synapses with skeletons, (c) aggregating synapses into clusters,  (d) classifying skeleton branches, and (e) detecting errors to guide proofreading.  (f) Errors are detected by a module based on a recurrent network that detects inconsistencies between an object's mask and its skeleton. 
}
\vspace{-10pt}
\end{figure*}
\vspace{3mm}\\
\noindent
The \textbf{Circuit Viewer}\label{sec:circuit_viewer}, shown in Figure~\ref{fig:userinterface} (b), displays the local circuit of a single cell in 3D in support of tasks \textbf{T1-T4}. Limiting the display to the local circuit of a single cell ensures that users can explore and examine cellular components without visual clutter, and large neurons can be visualized efficiently at interactive rates. The viewer renders cellular components (errors, skeletons, and synapses) in 3D and enables user interaction through simple point-and-click mouse interaction and surface selection.  We complement the user interaction with a third-person camera that attaches to object surfaces on selection. This type of camera is ideal for our application because it provides the user with a means for scrutinizing 3D bodies of cellular components from different angles. Finally, the viewer provides a set of tools, shown at the bottom of Figure~\ref{fig:userinterface} (b), to enable users to control the camera, toggle object visibility, and add custom annotations.
\vspace{3mm}\\
\noindent
The \textbf{Circuit Inspector}\label{sec:segment_editor},
shown in Figure~\ref{fig:userinterface} (c), is the most critical component of VICE to enable targeted proofreading with on-demand details.  The inspector attaches to the 3D body of a cellular component and enables users to examine and scrutinize the component with a global context from different angles or local context at the pixel level. In the global context, users can resolve errors by tagging surfaces of 3D objects, while at the pixel-level, users can edit the segmentation of an object by painting with a 3D brush on an in-place 2D cross-section plane. To enable inspection and to edit at the pixel-level across slices, the user can scroll the inspection volume with a slice navigator, shown to the right of Figure~\ref{fig:userinterface}(b). We link the 2D cross-section plane to a 2D cross-section view of the volume at scale, as shown in Figure~\ref{fig:userinterface}(d), to allow users to quickly determine whether a segmentation error is caused by an image artifact or some randomness in the EM data. This 3D-to-2D cross-section linkage helps the user to view the inspected-object relative to other objects in the volume and identify other errors in its vicinity.  On attachment, the inspector loads data and automatic segmentation on a small region centered on the attachment point. The small region is limited to $512\times512\times100$ pixels$^3$ to ensure efficient inspection of large neurons at interactive rates. This region may contain multiple structures depending on the dataset.

\subsection{Identifying Connectivity Errors}
In this section, we describe our workflow for identifying connectivity errors.  
We address four types of errors based on heuristics from our collaborators: broken neurite, merged neurites, disconnected neurite, and invalid branch, as shown in Figure~\ref{fig:strategies}. These errors only represent a subset of all possible errors.

\noindent
\textbf{Skeleton Construction}:\label{sec:skeletonization}
We extract the skeleton of a neuron for efficient visualization, error detection, and proofreading along cell branches (\textbf{T1}, \textbf{T2}).  We compute the skeleton of a cell using the Kimimaro skeletonization framework \cite{kimimaro}, as shown in Figure \ref{fig:strategies}(a).  Each skeleton is a tree-like structure consisting of branching center-lines and radii, which are represented with pixels. This representation is compact and memory-friendly and enables us to load large cells from dense reconstructions (\textbf{C1}).  We classify the skeleton branches according to neighboring synapses, as shown in Figure \ref{fig:strategies}(d). Branches where presynaptic contacts reside, are classified as axons, those with postsynaptic contacts are dendrites, and the rest are miscellaneous. These classifications are based on feedback from our collaborators and can be overridden by the user. The final skeleton is visualized as a hierarchy of interactive 3D lines.

\noindent
\textbf{Broken Neurites Detection}:
Broken neurites are commonly caused by artifacts in images or unexpected appearance changes that cause automation to make wrong segment predictions.  To detect these errors, we have implemented an error detection module, shown in Figure~\ref{fig:strategies}(f), that analyzes every object in a volume to flag errors (\textbf{T1}).  This module incorporates our previous recurrent neural network~\cite{gonda2021consistent}.  The module detects inconsistencies between the mask and the skeleton of a single object. The error detector takes neuron's skeleton and its synapses as input and retraces the neurite from the cell body to its endpoint.  Tracing is performed with a small sliding window of ten images.  On each trace, we sequentially analyze the predicted object mask and flag an error when a missing segment is detected, i.e., when the predicted object extends beyond the skeleton endpoints. 
\begin{figure*}[t!]
\begin{minipage}[b]{1.0\linewidth}
  \centering \centerline{\includegraphics[clip, trim=0.0cm 0.0cm 0.0cm 0.0cm, width=\textwidth]{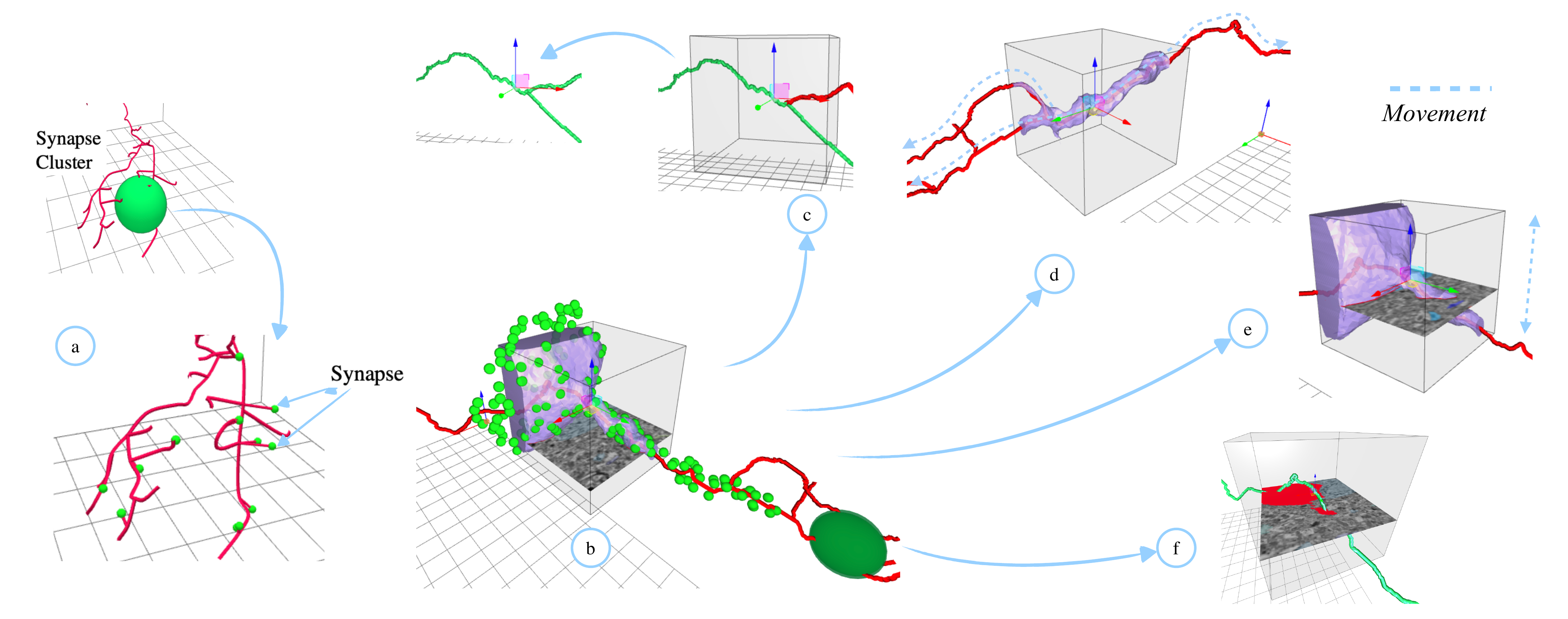}}
\end{minipage}
\caption{\label{fig:inspector}
(a) Synapse cluster inspection.  (b) The neural circuit inspector enables: (c) 3D error correction, (d) neurite inspection, (e) Hybrid 2D-3D segmentation inspection, and (f) pixel-level error correction via a 2D cross-section plane editor using a 3D paint brush. 
}
\vspace{-10pt}
\end{figure*} 

Unlike the original network, which extracts multiple object masks, we train the error detection network to extract single object masks from sequences of images.  We use images and ground-truth labels from the JWR and FIB25 datasets, described in Section~\ref{sec:domain_data}, to train our network. Our collaborators manually created the ground truth labels for the JWR dataset.  We optimized our network's parameters using the Adam optimizer with a learning rate of $10^{-6}$ and a batch size of 1 over 30 epochs.  The training was carried out on a single NVIDIA Titan X GPU with 12GB RAM for 36 hours. We conduct an evaluation of our error detector module on multiple datasets and report its performance in Section~\ref{sec:error_detector_eval}.

\noindent
\textbf{Disconnected Neurites Detection}:
The second source of errors is skeleton endpoints that terminate within a volume, typically for dendritic structures. These endpoints are expected to connect to other cells.  Therefore, we perform a radial check to detect synapses in the neighborhood of skeletal endpoints.  We generate an error ROI at the endpoint if no synapse is detected (\textbf{T1}).
\vspace{3mm}\\
\noindent
\textbf{Invalid Branch Detection}:
The third source of errors is neurites that branch in the wrong direction, as shown in Figure~\ref{fig:strategies}.  Most neurites typically diverge once they branch, like a tree.  An exception to the divergence is neurites that branch and then come back together, a rare case.  To detect branching errors, we compute a dot product between the forward vectors of the branching neurite and its stem at the branching point.  Branches that flow in the opposite direction of a stem are flagged as errors.

\noindent
\textbf{Merged Neurites Identification}:
Merge errors are challenging to detect automatically. Instead, we assist the visual search of merge errors by enabling the inspector tool to snap to a neurite and move along its path. This allows users to examine suspicious segments along the neurite's path and flag an error if necessary.  We enable this capability by allowing users to select cellular branches using the mouse.  To inspect the branch, users simply drag the inspector's handle or scroll with the mouse to move along the skeleton of the neurite, as shown in Figure~\ref{fig:inspector} (d).


\subsection{Resolving Connectivity Errors}\label{sec:correcting_segments}
To fully satisfy the task (\textbf{T1}), we visually expose errors in the browser by name and in the viewer as spheres. Our collaborators preferred a single visual encoding for all errors instead of distinguishing them by type. To resolve errors, we provide the inspector tool, shown in Figure~\ref{fig:inspector}, enabling users to examine errors from different angles to resolve them.  For broken neurites, users can tag their endpoints to reconnect them at the object level.  For merged neurites and invalid branches, users can mark surface points on the neurite to split it into two.  These errors can also be corrected by painting on the 2D cross-section plane of the inspector, as shown in Figure~\ref{fig:inspector} (c). Invalid errors can be deleted from the circuit browser using the delete key. All segmentation changes are saved as edits, separate from the original segmentation, to enable simple versioning and rollbacks.


\subsection{Optimizing Synapse Proofreading}
In this section, we describe our strategies to deal with the density of synapses (\textbf{C3}) to accelerate synapse validation (\textbf{T2}).

\noindent
\textbf{Synapse Clusters Formation}\label{sec:synapse_clustering} 
To enable proofreading of synapses at scale, we form clusters that group co-located synapses along neurite paths into non-overlapping sets that can be proofread jointly. Cluster formation is performed by traversing each neurite's skeleton path at equal intervals and aggregating synapses within proximity of a traversal point into clusters. The interval and proximity are based on an application-defined radius. The Euclidean distance between each synapse and the traversal point determines to which cluster the synapse belongs. For example, in Figure~\ref{fig:strategies}(b), cluster \textit{CL1} is formed first, followed by \textit{CL2,} then \textit{CL3}. Each synapse belongs to only one cluster, and the cluster's location is the average position of all synapses contained. These clusters make it possible to visualize dense synapses efficiently (\textbf{C3}).  The clusters are rendered as spheres, and when selected, the corresponding synapses are displayed, as depicted in Figure~\ref{fig:inspector} (a). 
\begin{figure*}[t!] 
\begin{minipage}[b]{1.0\linewidth}
  \centering \centerline{\includegraphics[clip, trim=0.0cm 9.50cm 0.0cm 9.0cm, width=\textwidth]{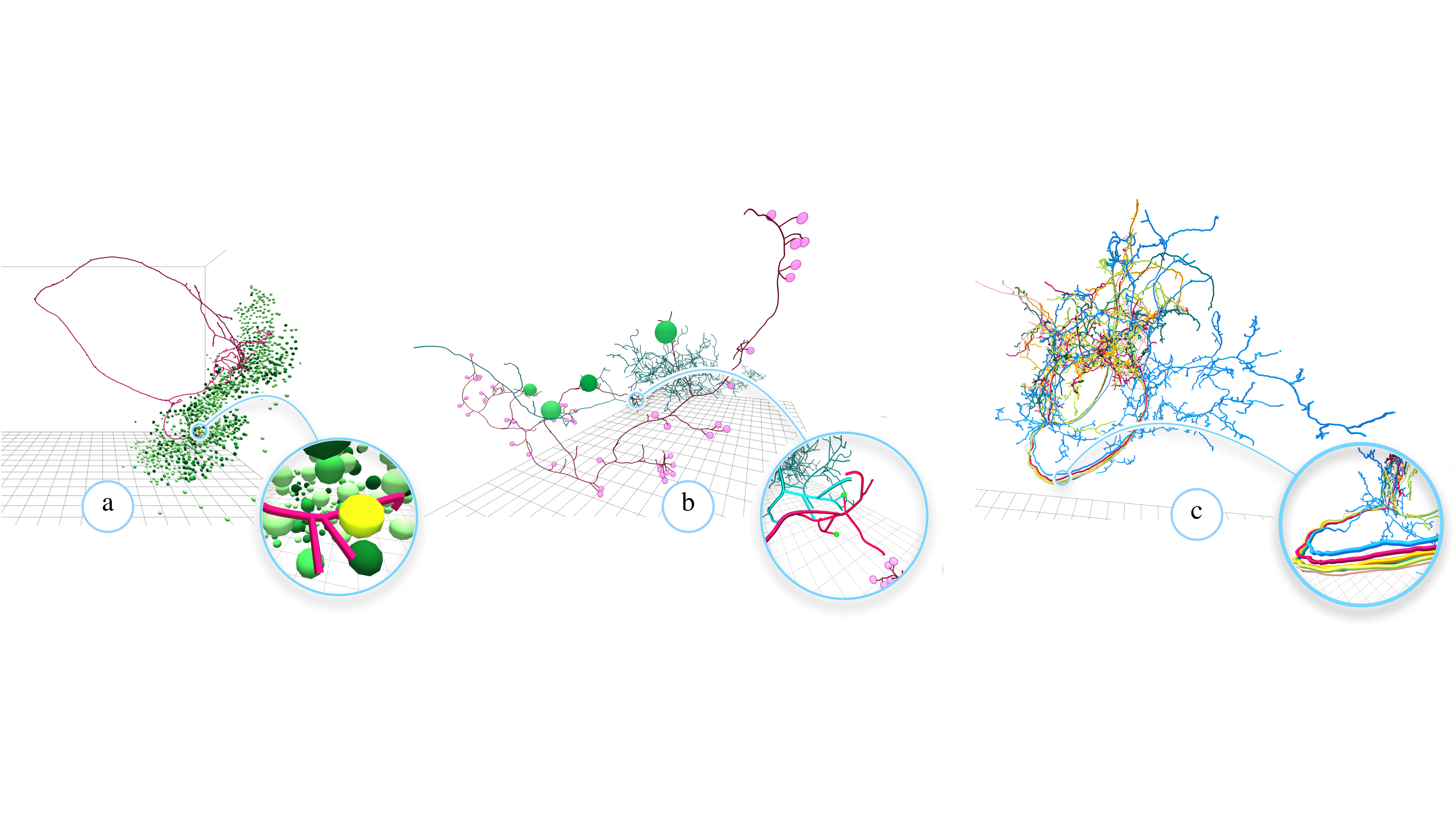}}
\end{minipage}
\caption{\label{fig:local_circuit}
(a) Global connectivity level showing the spatial distribution of cell bodies and synapse density. Local connectivity level showing (b) connectivity and cellular compartments, and (c) structures with similar morphological features.
}
\vspace{-10pt}
\end{figure*} 

\noindent
\textbf{Synapse Clusters Sorting}\label{sec:synapse_sorting}
To enable efficient proofreading of synapses along cell branches (\textbf{T2}), we sort the clusters spatially to synchronize the order in which they appear in the circuit browser to the viewer.  We perform sorting by ordering clusters based on their distance from the skeleton's root to the skeleton branches' endpoints. This ordering ensures that synapse clusters of each skeleton branch are contiguously positioned in the browser. As such, users can navigate the clusters sequentially to ensure completeness of one branch before proofreading the next, thus alleviating the mental demand incurred by switching between random synapse locations in VAST++ (\textbf{T2}).

\noindent
\textbf{Synapse Validation}
To validate synapses, we use the inspector tool to examine a cluster of synapses jointly, as shown in Figure~\ref{fig:inspector} (b).  This strategy enables the user to quickly perform bulk operations such as invalidating or classifying a group of synapses at once, thus dramatically reducing turnaround time. When a synapse cluster is selected, as shown in Figure~\ref{fig:inspector} (a), we display its corresponding synapses to enable examination in 3D or 2D with a 2D cross-section plane at a pixel level, as shown in Figure~\ref{fig:inspector} (b).  Users can also edit synapses with the 2D plane. Users can add synaptic elements at any position, connect or disconnect elements, or remove or move elements.  Finally, synapses can be classified in bulk by multi-selecting and assigning the corresponding class (\textbf{T4}).

\subsection{Examining Connectivity Graphs}
To enable exploratory analysis and targeted proofreading, we provide global and local exploration modes in the circuit browser. 

\noindent
At the \textbf{Global Connectivity} level, our collaborators can examine the distribution and the concentration of errors and synapses in a volume. Our collaborators often use the cell body as the starting point of proofreading and the reference point of analysis. Therefore, we encode the cell body as a sphere that is shaded based on the proofreading objective.  For proofreading neurites, we shade the cell bodies red to signify the number of errors, while for synapse validation, we shade them green to signify density, as shown in Figure~\ref{fig:local_circuit}(a).  Darker shades signify a greater number of errors or synapses. Our collaborators use these shading schemes to quickly examine areas of the circuit where proofreading resources are needed most (\textbf{T3}).

\noindent
At the \textbf{Local Connectivity} level, we visualize the compartments of a cell and its local circuit, as shown in Figures~\ref{fig:local_circuit} (b).  This level enables users to examine connection properties and the morphological structures on which presynaptic and postsynaptic elements reside.  The local level also enables users to carry-out targeted proofreading of cells with similar morphological features, as shown in Figure~\ref{fig:local_circuit} (c). Researchers often look for new prominent features or patterns in the data and pick cells or certain cellular compartments, such as a group of axons, for detailed analysis. Users can examine and isolate similar cells by grouping them into new sub-graphs and proofreading them separately (\textbf{T3}).
\subsection{Implementation}\label{sec:implementation}
We implemented VICE as a three-tiered architecture consisting of a web-based front-end, an application logic layer, and a central repository where segmentation revisions and graph data is saved.  The front-end is developed using HTML5 and the Three.Js library. The application logic layer is written in python using the Pytorch library. The webserver used Flask \cite{flask}. We implemented two rendering modes for 3D graph components. In \emph{proofreading mode}, we render 3D elements in low resolution without shaders to ensure maximum performance.  In \emph{presentation mode}, we render 3D elements with custom shaders to generate high-quality images and videos.  We further improve performance by excluding 3D elements of postsynaptic cells from ray-testing since only presynaptic elements are selectable in the viewer.

\section{Evaluation}
\subsection{Data}\label{sec:domain_data}
We use three large-scale connectomic datasets (volumes), described in Table \ref{datasets}.  The first dataset, JWR ($106\times106\times 93 $µm$^{3}$), is from our collaborators. The second dataset, FIB-25 ($36 \times 29 \times 69 $µm$^{3}$), and the third dataset, a reconstruction of the \textit{Drosophila melanogaste} fruit fly brain \cite{hemibrain} ($250 \times 250 \times 250 $µm$^{3}$), are from Janelia Research Campus.
\begin{table}[t!]
\centering
\begin{tabular}{l c c c} 
 \hline
 Name & Species & No. Neurons & No. Synapses \\ [0.5ex] 
 \hline\hline
 JWR & Rat &  $6$ & $10,203$ \\ 
 FIB25 & Fruit Fly &  $491$ & $63,258$ \\ 
 \textit{Drosophila} & Fly &  $25$K & $9.6$M \\ [1ex] 
 \hline
\end{tabular}
\caption{List of datasets used to develop VICE.}
\label{datasets}
\end{table}
\begin{table}[t!]
\centering
{\begin{tabular}{@{}lcc}
  \hline
 Dataset & Pre-Proofreading (ARI) & Post-Proofreading (ARI)  ~~ \\
\hline
JWR & 0.25 & 0.02 \\
FIB25 & 0.31 & 0.05 \\
  \hline
\end{tabular}}
\caption{Reported accuracy results in terms of ARI (lower is better), before and after proofreading with our error detection.
}
\label{tab:error_detector}
\end{table}

\subsection{Error Detector Evaluation}\label{sec:error_detector_eval}

We assess our error detector's accuracy with six neurons per dataset, each from the JWR and FIB25 datasets. We first detect errors, which a proofreader then corrects.  We then compare the corrected segmentation to the ground truth using the Adaptive Rand Index (ARI) \cite{randindex}, a metric commonly used for connectomics data.  The ARI measures the similarity between two data clusters. The error is defined as one minus the Rand index's maximal F-score, where a lower score corresponds to better segmentation.  In Table~\ref{tab:error_detector}, the ARI results show that our system can identify suspicious locations that lead to better segmentation when proofread by a human expert. Our results do not account for merged neurites, which require visual search by a human expert.

\subsection{User Study}
This section evaluates the effectiveness, efficiency, and satisfaction of using our method on our collaborators' use cases. To that end, we report on a user study that focuses on four questions:
\begin{enumerate}[label=\textbf{Q\arabic*}]
\item Can users with expertise in proofreading EM data \textit{learn to use the proposed system} to perform proofreading tasks?
\item Does the \textit{unified interface} of VICE improve usability over the use of multiple applications in VAST++?
\item While we know VAST++ demands extensive time to inspect image data and correct errors, \textit{can we quantify time spent} proofreading neurites and synapses?
\item Finally, we are interested in users' \textit{subjective preferences} between VICE and VAST++.
\end{enumerate}

\subsubsection{Proofreading Tasks}
We selected eight tasks that represent our collaborator's most common proofreading tasks.  

\begin{enumerate}[label=\textbf{S\arabic*}]
\item Identify and group cells with similar morphological features. 
\item Mark endpoints of \textit{broken neurites} to reconnect them.
\item Mark surface point of \textit{merged neurites} to split them apart.
\item Inspect endpoints of \textit{disconnected neurites} for synapses.
\item Assign labels to groups of neurites to \textit{classify} them.
\item Inspect and validate all synapses.
\item Assign labels to groups of synapses to \textit{classify} them.
\item Inspect connectivity pathways to partner cells.
\end{enumerate}

\subsubsection{Participants and Setup}
We recruited five participants from the Center for Brain Science at Harvard University, consisting of one neuroscientist (a co-author) and four trained proofreaders. We used a low number of experts instead of many novice users to evaluate the usability and performance improvements of our system compared to the previous workflow. All participants were highly familiar with VAST++ but had never used VICE. The investigator reviewed the study plans with participants and instructed them to complete the study within 20 hours per system.  Participants first completed all tasks in VICE before starting the same tasks in VAST++.   For each task, users had to complete a test example first, ensuring they understood the task correctly.  Then, they completed the eight proofreading tasks (\textbf{S1-S8}).  Participants started a task by selecting a cell and recording the starting time of the task.  Participants then performed the task and recorded their completion time.  Participants then answered post-task questions and could choose to have a break before starting the next task. The investigator observed the study over a video conference.   After completing the tasks, participants completed a post-study questionnaire and gave feedback to the investigator.

\subsubsection{Results}
\begin{figure}[t]
\begin{minipage}[b]{1.0\linewidth}
  \centering \centerline{\includegraphics[clip, trim=5.50cm 1.00cm 5.50cm 1.50cm, width=\textwidth]{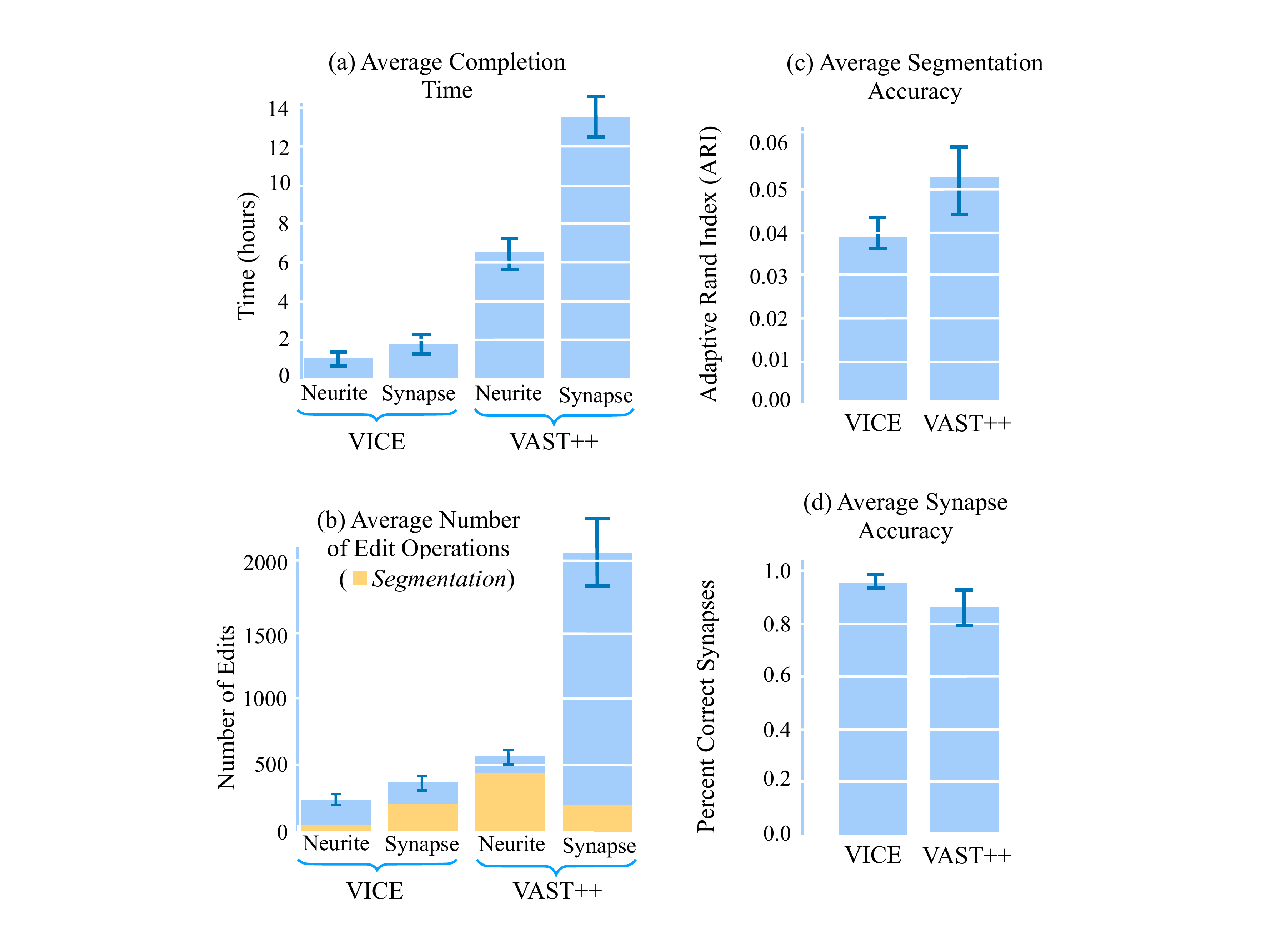}}
\end{minipage}
\caption{\label{fig:proofreading_results}
Completion time, number of edit operations, and accuracy for proofreading neurites and synapses. Lower ARI is better.
}
\vspace{-10pt}
\end{figure}

In Figure~\ref{fig:proofreading_results}, we report the performance comparison of VICE and VAST++ in terms of accuracy, completion time, and the number of edit operations required to proofread a single neuron fully.  We counted the number of edit operations automatically when participants modified the graph with add, update, or delete operations.  Six neurons were used in the study consisting of 2500 synapses on average. Due to the limited number of participants, we do not report on statistical significance.

\noindent
\textbf{Completion Time}\\
Since the distribution of completion time can vary by participants, we report the average time of all participants for proofreading the same neuron in Figure~\ref{fig:proofreading_results} (a).  A pairwise comparison indicates that our approach significantly outperforms the previous workflow.  On average, participants spend 1 hour (SE=0.09) and 6.5 hours (SE=1.3) proofreading neurites in VICE and VAST++, respectively.  For synapse validation, participants spend on average 2 hours (SE=0.64) and 13.5 hours (SE=1.1) in VICE and VAST++, respectively. The completion time results confirm that the error ROIs, synapse clusters, and automatic neurites' classification accelerate proofreading tasks. Particularly for synapse validation, participants visit synapse locations randomly in VAST++, leading to redundant visual inspections (\textbf{Q3}). The completion time results also confirm that centralizing proofreading tasks in a common interface could alleviate users' mental demand previously incurred by using multiple applications in VAST++.

\noindent
\textbf{Number of Edit Operations}\\
In Figure~\ref{fig:proofreading_results} (b), we report the average number of edit operations initiated by participants to proofread the same neuron fully.  As the results demonstrate, our approach requires fewer edits than VAST++. On average, participants initiated 200 (SE=15) and 620 edits (SE=21) proofreading neurites in VICE and VAST++, respectively.  For synapse validation, participants initiated 380 (SE=37) and 2200 edits (SE=156) in VICE and VAST++, respectively.  We highlight the number of segmentation edits in yellow.  In VAST++, most time is spent visually inspecting data and entering information into a spreadsheet.  Whereas in VICE, the clustering of synapses and the use of bulk operations drastically reduces classification tasks.  The results also confirm that resolving neurite errors at the object level (VICE) is faster than at the pixel level (VAST++).

\noindent
\textbf{Accuracy}\\
In Figure~\ref{fig:proofreading_results} (c), we report the mean accuracy for segmentation changes from all participants in terms of the ARI metric described in Section~\ref{sec:error_detector_eval}.  On average, the participant's segmentation accuracy is similar for both methods, with 0.038 (SE=0.004) and 0.052 (SE=0.012) for VICE and VAST++, respectively.   For synapse validation, we report the average percentage of correctly identified synapses compared to the ground truth, as shown in Figure~\ref{fig:proofreading_results} (d).  On average, participant's synapse accuracy was 0.97 (SE=0.01) and 0.84 (SE=0.06) for VICE and VAST++ respectively. We hypothesize the lower accuracy for synapse validation in VAST++ is partly due to the time limit imposed on the study and the random inspection of synapse in VAST++.

\noindent
\textbf{Subjective User Feedback}\\
After the user study, we asked users to rate VICE and VAST++ on a 0 (very bad) - 5 (very good) Likert scale and provide additional comments.  We used the following questions:
\begin{itemize}
    \item How confident did you feel with the technique? (Confidence)
    \item How easy was the user interface to learn? (Learnability),
    \item How clutter-free was the technique? (Clutter-free),
    \item Does the system support proofreading tasks? (Completeness)
    \item How would you rate the system overall? (Overall)
\end{itemize}

\begin{figure}[t]
\begin{minipage}[b]{1.0\linewidth}
  \centering \centerline{\includegraphics[clip, trim=8.0cm 9.0cm 8.0cm 9.0cm, width=\textwidth]{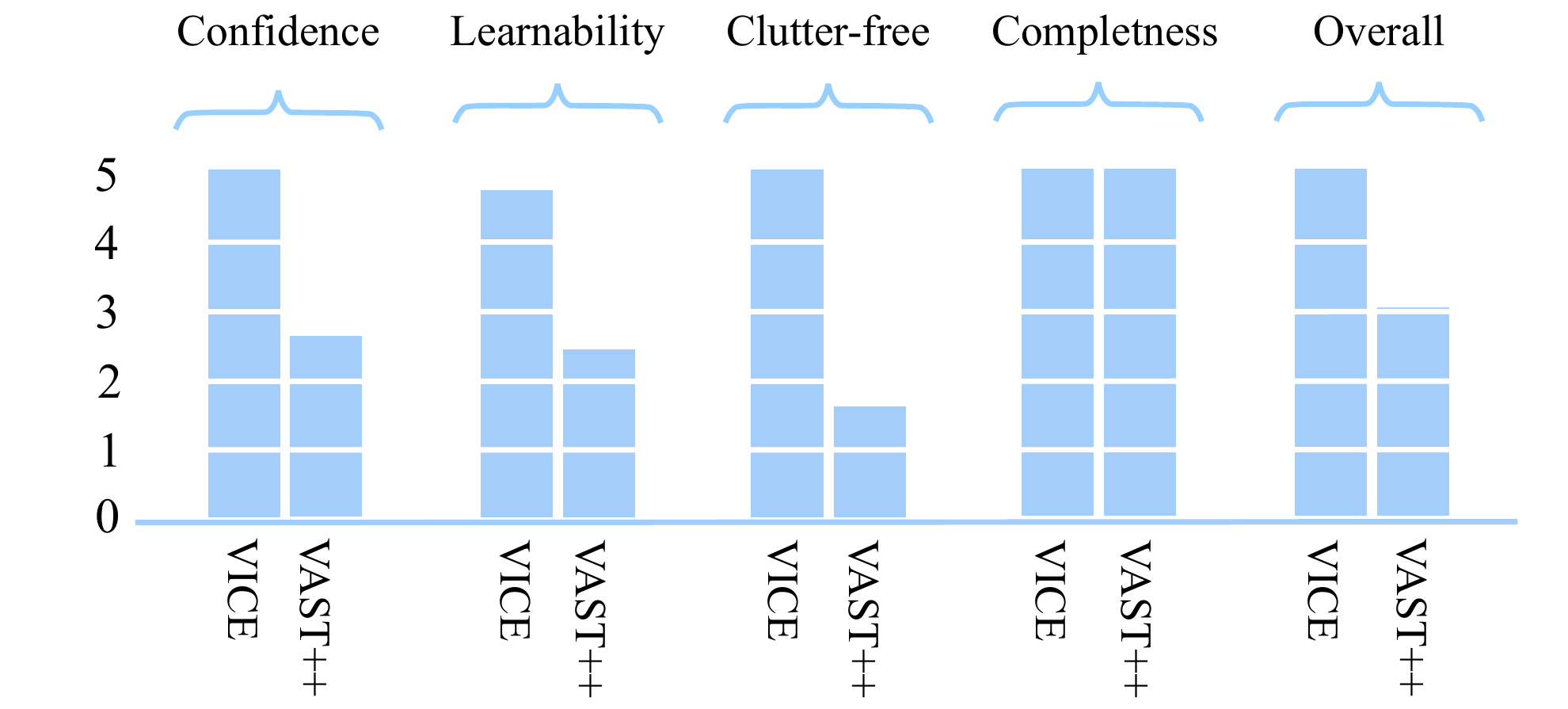}}
\end{minipage}
\caption{\label{fig:subjective_results}
Reported user subjective preferences.
}
\vspace{-10pt}
\end{figure} 
Figure~\ref{fig:subjective_results} shows the ratings provided by participants to our questions.  We now report salient insights from participant's answers.  Mostly, participants agreed on the clutter-free and overall assessment of the systems but diverged on confidence and learnability.  On completeness, participants indicated that both systems provided the necessary tools for accomplishing proofreading tasks.

(\textbf{Q4}) In terms of learnability, participants' answers indicated that they considered VICE easier to learn even though a user guide was not provided.  We think this is because of the clean interface design of VICE. In contrast, VAST++ requires participants to learn three separate tools.  On clutter-free, we expected participants to rate VAST++ lower (especially regarding 3D inspection) since it requires participants to disable cells to de-clutter views manually.  The most interesting result is about confidence in completing a task with each approach. While we expected participants to have high confidence in VICE, we didn't expect a lower rating for VAST++. This may be partially explained by ratings in learnability since VAST++ was generally found the most difficult to understand by participants since it requires learning three separate tools. 

\section{Discussion and Future Work}
The results of our study indicate that VICE is an understandable tool for new users (\textbf{Q1}), and has some advantages over VAST++: it allows for similar proofreading tasks but faster completion time and is generally preferred by participants, particularly for the unified interface and the inspection tool. The analysis of the number of editing operations pointed to the high degree of manual classification and documentation inherent in the VAST++ approach. In our VICE approach, the synapse clustering and automatic classification of neurites allowed participants to perform bulk documentation and classification, thus significantly reducing the number of edits.  The results also confirmed that VAST++ was less effective (\textbf{Q2}) in terms of task completion. Participants noted that they often took breaks when using VAST++ due to exhaustion from context-switching between applications and random inspection of synapses.   



To conclude our study: our proposed approach proved to be most effective in reducing the time to perform proofreading tasks (\textbf{Q3}). However, it is essential to note that our approach requires initial learning to exploit its most performant features, such as the in-place editing in 3D and the bulk classification of cellular components (\textbf{Q1}). VICE integrates several tools into one, which helps reduce the mental overload otherwise incurred by VAST++.

\noindent
\textbf{Future Studies}\\
While we tested only six neurons in our study, the local connectivity pathways and pathways-preserving structures are what we wanted to explore in this study.  Future studies should investigate connectivity beyond the local circuit of a cell and especially focus on higher-level tasks such as identifying repeated patterns (motifs).  In general, higher-level exploration tasks such as comparing local circuits, understanding the evolution of a circuit over time are under-explored.  We also envision further studies on the classification of cellular components based on biological function.  These classifications could be derived from heuristics and automated to help further reduce manual efforts.  

\section{Conclusion}
We conclude that our proposed local-circuit proofreading approach assisted with automated detection of likely errors  presents an effective approach to reducing human effort in proofreading neuronal pathways. With increasing advances in segmentation algorithms, automation can capture most neuronal structures and synapses, and with our approach, users only need to focus on connectivity-preserving errors.  In providing a working implementation, we observed usage patterns and understood the potentials and drawbacks of our approach.   We found that our approach works best when a high degree of synapses is known, enabling bulk proofreading and automatic classification.  Our single-cell approach also reduces visual clutter by rendering only the local circuit of a cell. We aim to contribute to the neuroscience community and make the application available here: \url{https://fegonda.github.io/vice}.

\section{Acknowledgments}
We wish to thank all participants from our user studies who helped us evaluate VICE. This work was partially supported by NSF NCS-FO grant IIS-1835231.

\bibliographystyle{eg-alpha-doi}
\bibliography{root}

\newcommand{\etalchar}[1]{$^{#1}$}
\begin{thebibliography}{\uppercase{BAAK{\etalchar{*}}13}}

\bibitem[AABH{\etalchar{*}}15]{awami_neuroblocks_2015}
\textsc{Al-Awami A., Beyer J., Haehn D., Kasthuri N., Lichtman J., Pfister H.,
  Hadwiger M.}:
\newblock Neuroblocks - visual tracking of segmentation and proofreading for
  large connectomics projects.
\newblock \emph{IEEE Transactions on Visualization and Computer Graphics
  (Proceedings IEEE SciVis 2015) 22}, 1 (2015), 738--746.

\bibitem[ABS{\etalchar{*}}14]{neurolines}
\textsc{Al{-}Awami A.~K., Beyer J., Strobelt H., Kasthuri N., Lichtman J.~W.,
  Pfister H., Hadwiger M.}:
\newblock {NeuroLines}: {A} subway map metaphor for visualizing nanoscale
  neuronal connectivity.
\newblock \emph{{IEEE} Transactions on Visualization and Computer Graphics 20},
  12 (2014), 2369--2378.

\bibitem[BAAK{\etalchar{*}}13]{ConnectomeExplorer}
\textsc{Beyer J., Al-Awami A., Kasthuri N., Lichtman J.~W., Pfister H.,
  Hadwiger M.}:
\newblock {ConnectomeExplorer}: Query-guided visual analysis of large
  volumetric neuroscience data.
\newblock \emph{IEEE Transactions on Visualization and Computer Graphics 19},
  12 (Dec. 2013), 2868--2877.
\newblock URL: \url{http://dx.doi.org/10.1109/TVCG.2013.142}, \href
  {https://doi.org/10.1109/TVCG.2013.142} {\path{doi:10.1109/TVCG.2013.142}}.

\bibitem[BLK{\etalchar{*}}11]{netanatomy}
\textsc{Bock D., Lee W.-C., Kerlin A., Andermann M., Hood G., Wetzel A.,
  Yurgenson S., Soucy E., Kim H., Reid R.~C.}:
\newblock Network anatomy and in vivo physiology of visual cortical neurons.
\newblock \emph{Nature 471} (03 2011), 177--82.
\newblock \href {https://doi.org/10.1038/nature09802}
  {\path{doi:10.1038/nature09802}}.

\bibitem[BSL18]{vast}
\textsc{Berger D.~R., Seung H.~S., Lichtman J.~W.}:
\newblock Vast (volume annotation and segmentation tool): Efficient manual and
  semi-automatic labeling of large {3D} image stacks.
\newblock \emph{Frontiers in Neural Circuits 12} (2018), 88.
\newblock URL:
  \url{https://www.frontiersin.org/article/10.3389/fncir.2018.00088}, \href
  {https://doi.org/10.3389/fncir.2018.00088}
  {\path{doi:10.3389/fncir.2018.00088}}.

\bibitem[CDU{\etalchar{*}}20]{neuprint}
\textsc{Clements J., Dolafi T., Umayam L., Neubarth N.~L., Berg S., Scheffer
  L.~K., Plaza S.~M.}:
\newblock {neuPrint}: Analysis tools for em connectomics.
\newblock \emph{bioRxiv} (2020).
\newblock URL:
  \url{https://www.biorxiv.org/content/early/2020/01/17/2020.01.16.909465},
  \href
  {http://arxiv.org/abs/https://www.biorxiv.org/content/early/2020/01/17/2020.01.16.909465.full.pdf}
  {\path{arXiv:https://www.biorxiv.org/content/early/2020/01/17/2020.01.16.909465.full.pdf}},
  \href {https://doi.org/10.1101/2020.01.16.909465}
  {\path{doi:10.1101/2020.01.16.909465}}.

\bibitem[FTG{\etalchar{*}}19]{Funke2019LargeSI}
\textsc{Funke J., Tschopp F.~D., Grisaitis W., Sheridan A., Singh C., Saalfeld
  S., Turaga S.~C.}:
\newblock Large scale image segmentation with structured loss based deep
  learning for connectome reconstruction.
\newblock \emph{IEEE Transactions on Pattern Analysis and Machine Intelligence
  41} (2019), 1669--1680.

\bibitem[GWP21]{gonda2021consistent}
\textsc{Gonda F., Wei D., Pfister H.}:
\newblock Consistent recurrent neural networks for 3d neuron segmentation,
  2021.
\newblock \href {http://arxiv.org/abs/2102.01021} {\path{arXiv:2102.01021}}.

\bibitem[HHM{\etalchar{*}}17]{daniel2017scalable}
\textsc{Haehn D., Hoffer J., Matejek B., Suissa-Peleg A., Al-Awami A.~K.,
  Kamentsky L., Gonda F., Meng E., Zhang W., Schalek R., Wilson A., Parag T.,
  Beyer J., Kaynig V., Jones T.~R., Tompkin J., Hadwiger M., Lichtman J.~W.,
  Pfister H.}:
\newblock Scalable interactive visualization for connectomics.
\newblock \emph{Informatics 4}, 3 (2017).
\newblock \href {https://doi.org/10.3390/informatics4030029}
  {\path{doi:10.3390/informatics4030029}}.

\bibitem[HKBR{\etalchar{*}}14]{dojo}
\textsc{Haehn D., Knowles-Barley S., Roberts M., Beyer J., Kasthuri N.,
  Lichtman J.~W., Pfister H.}:
\newblock Design and evaluation of interactive proofreading tools for
  connectomics.
\newblock \emph{IEEE Transactions on Visualization and Computer Graphics 20},
  12 (2014), 2466--2475.
\newblock URL: \url{http://rhoana.org/dojo/}.

\bibitem[HKT{\etalchar{*}}17]{guidedproofreading}
\textsc{Haehn D., Kaynig V., Tompkin J., Lichtman J.~W., Pfister H.}:
\newblock Guided proofreading of automatic segmentations for connectomics.
\newblock \emph{CoRR abs/1704.00848} (2017).
\newblock URL: \url{http://arxiv.org/abs/1704.00848}, \href
  {http://arxiv.org/abs/1704.00848} {\path{arXiv:1704.00848}}.

\bibitem[HSP16]{synapsepred}
\textsc{Huang G.~B., Scheffer L.~K., Plaza S.~M.}:
\newblock Fully-automatic synapse prediction and validation on a large data
  set.
\newblock \emph{CoRR abs/1604.03075} (2016).
\newblock URL: \url{http://arxiv.org/abs/1604.03075}, \href
  {http://arxiv.org/abs/1604.03075} {\path{arXiv:1604.03075}}.

\bibitem[{ica}20]{threejs}
\textsc{{icardo Cabello}}:
\newblock {Three.js}, 2020.
\newblock \url{https://github.com/mrdoob/three.js}, Last accessed on
  2020-03-09.

\bibitem[JKL{\etalchar{*}}18]{floodfilling}
\textsc{Januszewski M., Kornfeld J., Li P.~H., Pope A., Blakely T., Lindsey L.,
  Maitin-Shepard J.~B., Tyka M., Denk W., Jain V.}:
\newblock High-precision automated reconstruction of neurons with flood-filling
  networks.
\newblock \emph{Nature Methods 15} (2018), 605--610.
\newblock URL: \url{https://www.nature.com/articles/s41592-018-0049-4}.

\bibitem[neu]{neuroglancer}
Neuroglancer: {WebGL}-based viewer for volumetric data.
\newblock \url{https://github.com/ google/neuroglancer}.
\newblock Accessed: 2019-03-09.

\bibitem[Pal]{flask}
\textsc{Pallets}:
\newblock Flask.
\newblock \url{https://flask.palletsprojects.com/en/1.1.x/}.
\newblock Accessed: 2020-01-03.

\bibitem[PF18]{AnalyzingImage}
\textsc{Plaza S.~M., Funke J.}:
\newblock Analyzing image segmentation for connectomics.
\newblock \emph{Frontiers in Neural Circuits 12} (2018), 102.
\newblock URL:
  \url{https://www.frontiersin.org/article/10.3389/fncir.2018.00102}, \href
  {https://doi.org/10.3389/fncir.2018.00102}
  {\path{doi:10.3389/fncir.2018.00102}}.

\bibitem[PKB{\etalchar{*}}12]{visconnectomics}
\textsc{Pfister H., Kaynig V., Botha C.~P., Bruckner S., Dercksen V.~J., Hege
  H., Roerdink J. B. T.~M.}:
\newblock Visualization in connectomics.
\newblock \emph{CoRR abs/1206.1428} (2012).
\newblock URL: \url{http://arxiv.org/abs/1206.1428}, \href
  {http://arxiv.org/abs/1206.1428} {\path{arXiv:1206.1428}}.

\bibitem[SCHT09]{catmaid}
\textsc{Saalfeld S., Cardona A., Hartenstein V., Tomancak P.}:
\newblock {CATMAID: collaborative annotation toolkit for massive amounts of
  image data}.
\newblock \emph{Bioinformatics 25}, 15 (04 2009), 1984--1986.
\newblock URL: \url{https://doi.org/10.1093/bioinformatics/btp266}, \href
  {http://arxiv.org/abs/https://academic.oup.com/bioinformatics/article-pdf/25/15/1984/555362/btp266.pdf}
  {\path{arXiv:https://academic.oup.com/bioinformatics/article-pdf/25/15/1984/555362/btp266.pdf}},
  \href {https://doi.org/10.1093/bioinformatics/btp266}
  {\path{doi:10.1093/bioinformatics/btp266}}.

\bibitem[Seu12]{connectome}
\textsc{Seung S.}:
\newblock \emph{Connectome: How the Brain’s Wiring Makes Us Who We Are}.
\newblock HMH, 2012.

\bibitem[Seu20]{eyewire}
\textsc{Seung S.}:
\newblock \emph{eyewire}, 2012 (accessed April 16, 2020).
\newblock URL: \url{http://eyewire.org}.

\bibitem[Shn96]{eyeshaveit}
\textsc{Shneiderman B.}:
\newblock The eyes have it: A task by data type taxonomy for information
  visualizations.
\newblock In \emph{Proceedings of the 1996 IEEE Symposium on Visual Languages}
  (USA, 1996), VL '96, IEEE Computer Society, p.~336.

\bibitem[Sil]{kimimaro}
\textsc{Silversmith W.~M.}:
\newblock Kimimaro.
\newblock \url{https://github.com/seung-lab/kimimaro}.
\newblock Accessed: 2020-03-09.

\bibitem[SLK{\etalchar{*}}16]{richard2016imaging}
\textsc{Schalek R., Lee D., Kasthuri N., Suissa-Peleg A., Jones T.~R., Kaynig
  V., Haehn D., Pfister H., Cox D., Lichtman J.~W.}:
\newblock Imaging a 1 mm$^3$ volume of rat cortex using a multibeam sem.
\newblock In \emph{Microscopy and Microanalysis} (26 July 2016), vol.~22,
  Cambridge Univ Press, pp.~582--583.

\bibitem[TXL{\etalchar{*}}15]{Takemura13711}
\textsc{Takemura S.-y., Xu C.~S., Lu Z., Rivlin P.~K., Parag T., Olbris D.~J.,
  others.}:
\newblock Synaptic circuits and their variations within different columns in
  the visual system of \textit{Drosophila}.
\newblock \emph{Proceedings of the National Academy of Sciences 112}, 44
  (2015), 13711--13716.
\newblock URL: \url{https://www.pnas.org/content/112/44/13711}, \href
  {http://arxiv.org/abs/https://www.pnas.org/content/112/44/13711.full.pdf}
  {\path{arXiv:https://www.pnas.org/content/112/44/13711.full.pdf}}, \href
  {https://doi.org/10.1073/pnas.1509820112}
  {\path{doi:10.1073/pnas.1509820112}}.

\bibitem[UPH07]{randindex}
\textsc{Unnikrishnan R., Pantofaru C., Hebert M.}:
\newblock Toward objective evaluation of image segmentation algorithms.
\newblock \emph{IEEE Transactions on Pattern Analysis and Machine Intelligence
  29}, 6 (June 2007), 929--944.
\newblock \href {https://doi.org/10.1109/TPAMI.2007.1046}
  {\path{doi:10.1109/TPAMI.2007.1046}}.

\bibitem[{Vas}20]{3dforcegraph}
\textsc{{Vasco Asturiano}}:
\newblock {3D} force directed graph, 2020.
\newblock \url{https://github.com/vasturiano/3d-force-graph}, Last accessed on
  2020-03-09.

\bibitem[WGM{\etalchar{*}}16]{zebrafinch}
\textsc{Wanner A.~A., Genoud C., Masudi T., Siksou L., Friedrich R.~W.}:
\newblock Dense {EM-based} reconstruction of the interglomerular projectome in
  the zebrafish olfactory bulb.
\newblock \emph{Nature Neuroscience 19} (2016), 816--825.

\bibitem[WSTB86]{celagan}
\textsc{White J., Southgate E., Thomson J.~N., Brenner S.}:
\newblock The structure of the nervous system of the nematode \textit{C.
  elegans}.
\newblock \emph{Philosophical transactions Royal Society London 314} (1986),
  1--340.

\bibitem[XJL{\etalchar{*}}20]{hemibrain}
\textsc{Xu C.~S., Januszewski M., Lu Z., Takemura S.-y., Hayworth K.~J., Huang
  G., Shinomiya K., et~al.}:
\newblock A connectome of the adult \textit{Drosophila} central brain.
\newblock \emph{bioRxiv} (2020).
\newblock \href
  {http://arxiv.org/abs/https://www.biorxiv.org/content/early/2020/01/21/2020.01.21.911859.full.pdf}
  {\path{arXiv:https://www.biorxiv.org/content/early/2020/01/21/2020.01.21.911859.full.pdf}},
  \href {https://doi.org/10.1101/2020.01.21.911859}
  {\path{doi:10.1101/2020.01.21.911859}}.

\bibitem[ZOYP18]{neutu}
\textsc{Zhao T., Olbris D.~J., Yu Y., Plaza S.~M.}:
\newblock Neutu: Software for collaborative, large-scale, segmentation-based
  connectome reconstruction.
\newblock \emph{Frontiers in Neural Circuits 12} (2018), 101.

\end{thebibliography}




\end{document}